\documentclass{article}

\usepackage{template/spconf}

\usepackage{template/macro}
\usepackage[T1]{fontenc}
\usepackage[utf8]{inputenc}
\usepackage[english]{babel}

\usepackage{times}

\PassOptionsToPackage{hyphens}{url}
\usepackage{url}
\usepackage{hyperref}

\usepackage{graphicx}
\usepackage{caption}
\usepackage{tikz}
\usepackage{pgfplots}
\pgfplotsset{compat=1.17}

\usepackage{array}
\usepackage{booktabs}
\usepackage{microtype}
\usepackage{xcolor}
\usepackage{enumitem}
\setitemize{itemsep=0pt,parsep=0pt,topsep=0pt}
\setenumerate{itemsep=0pt,parsep=0pt,topsep=0pt}
\usepackage[ruled,vlined]{algorithm2e}

\renewcommand\paragraph[1]{\medskip\noindent{\bf #1}}

\title{Touring sampling with pushforward maps}

\twoauthors
{Vivien Cabannes}
{Meta AI, FAIR Labs\\
New York, USA}
{Charles Arnal}
{INRIA, Datashape team\\
Sophia-Antipolis, France}

\begin{document}

\maketitle

\begin{abstract}
    
The number of sampling methods could be daunting for a practitioner looking to cast powerful machine learning methods to their specific problem.
This paper takes a theoretical stance to review and organize many sampling approaches in the ``generative modeling'' setting, where one wants to generate new data that are similar to some training examples.
By revealing links between existing methods, it might prove useful to overcome some of the current challenges in sampling with diffusion models, such as long inference time due to diffusion simulation, or the lack of diversity in generated samples.

\end{abstract}


\section{Introduction}

Generative AI has received much attention recently.
It consists in generating realistic data, could it be drugs to cure disease \cite{GenAI1}, award-winning images \cite{GenAI2}, or chat bots \cite{GenAI3}.
McKinsey recently estimated that it could boost the global economy by up to \$4.4 trillion annually \cite{GenAI0}. 
Recent advances in generative AI are due to machine learning, where, rather than complex engineering rules, a machine learns to generate new data that resemble training samples.
Several generic\footnote{We let aside specific perspectives such as next token prediction for LLMs.} perspectives have been suggested to do so, ranging from variational auto-encoder, generative adversarial networks (GANs), stochastic diffusion models, or flow models.
On the one hand, a GAN provides fast inference but is hard to train; on the other hand diffusion models are easy to train but are slow to generate new samples.
Could we have the best of both worlds?

The crux of this paper is to tour the sampling literature to explore and organize principled ways to learn deterministic mappings between distributions, which lead to fast inference algorithms.
We distinguish three different perspectives, one based on test functions, one based on density likeliness, and one based on probability flows.

\paragraph{Setting \& Notations.}
This paper considers $\X \subset \R^d$ endowed with a target distribution $\mu_1 \in \prob\cX$, that generates samples $X \sim \mu_1$. 
Here, $\prob A$ denotes the set of probability measures over a space $A$.
The goal is to parameterize $\mu_1$ from a user-specified initial distribution $\nu \in \prob\cZ$ for some ``sampling'' space $\cZ$, with a deterministic pushforward map $\phi:\cZ\to\cX$.
In other terms, we want to learn $\phi$ such that one can cast a sample $Z\sim\nu$, which is easy to generate, as a sample $X=\phi(Z)\sim\mu_1$ following the desired distribution $\mu_1$.
Formally, we want $\nu(\phi^{-1}(A)) = \mu(A)$ for any measurable set $A\subset \cX$, which reads $\phi_\#\nu = \mu_1$ when written with pushforward notation.
In order to benefit from the extensive literature on transport maps between measures on $\X$, we will eventually lift $\nu$ into $\mu_0 \in \prob\cX$ through some canonical embedding $\xi:\cZ\to\cX$ such that $\xi_\# \nu = \mu_0$.
The learning of $\phi$ is reparameterized as the learning of $\psi:\cX\to\cX$ that maps $\mu_0$ to $\mu_1$, i.e. $\psi_\#\mu_0 = \mu_1$, with $\phi = \psi \circ \xi$.

\section{Statistics Matching Objectives}\label{sec:functions_matching}
The first approach we will consider leverages the duality bracket, defined for any  measurable function $f:\X\to\R$ and measure $\mu \in \cM_\X$ over $\X$ as
\begin{equation}
    \bscap{f}{\mu} := \int_{\X} f(x) \mu(\diff x) =: \E_{X\sim\mu}[f(X)].
\end{equation}
When $\mu$ is parameterized as $\mu=\phi_\#\nu$, this corresponds to
\[
    \bscap{f}{\phi_\#\nu} = \E_{Z\sim\nu}[f(\phi(Z))].
\]
To make sure that $\mu = \mu_1$, one can make sure that the actions of both measures are the same against any test functions (or equivalently that the action of the signed measure $\mu - \mu_1$ is null everywhere).

\paragraph{Maximum formulation.}
Introducing $\cF \subset \R^\X$ a space of test functions, one can look for the variational objective defined by the worse function
\begin{equation}
\label{eq:d-max}
    D_\infty(\mu, \mu_1; \cF) := \sup_{f \in \cF} \bscap{f}{\mu - \mu_1}.
\end{equation}
Those measures are known as integral probability metrics \cite{StatMax0}.

When $\cF$ is the unit ball $\cB_k$ of a reproducing kernel Hilbert space with kernel $k:\X\times\X\to\R$, \eqref{eq:d-max} becomes the maximum mean discrepancy, or MMD \cite{MMD0}
\[
  \label{eq:mmd}
    D_\infty^2(\mu,\mu_1; \cB_k) = \E[k(X, X') + k(Y, Y') - 2 k(X, Y)],
\]
where $X$, $X'$ are independent variables distributed according to $\mu$ and $Y$, $Y'$ follows $\mu_1$.
When $\cF$ is taken as all the functions from $\X$ to $[0, 1]$, it becomes the total variation defined as 
\[
    D_\infty(\mu, \mu_1; [0,1]^\X) = \max_{A\subset \X} \abs{\mu(A) - \mu_1(A)}.
\]
When $f$, the argument of the maximum, is learned with an adversarial neural network, this approach identifies to GANs. 
And when $\cF$ is taken as $\cC^{0,1}_1$ the set of all contractions, i.e., functions that are $1$-Lipschitz continuous, \eqref{eq:d-max} becomes the $W^1$-Wasserstein metric, which has motivated W-GANs \cite{WGAN}.

\paragraph{Guarantees.}
From those observations, one can deduce that when the span of $\cF$ is dense in $\cC^{0,1}_1$ (with the $W_1^*$-topology), $D_\infty$ is a distance.
As a consequence, the function $\mu \mapsto D_\infty(\mu, \mu_1)$ would have a unique local minimum, achieved for $\mu = \mu_1$, reachable through gradient descent.
In practice, the descent could take place with respect to the parameter $\theta$ which parameterizes the map $\phi_\theta$, itself parameterizing $\mu = (\phi_\theta)_\#\nu$.
Eventually, the space $\cF$ may be chosen to ensure the consistency of the loss $\mu \mapsto D_\infty(\mu,\mu_1)$ for the sole target $\mu_1$.\footnote{
	E.g., $\cF$ could be chosen as the image of a Stein operator \cite{Stein2}.
}

\paragraph{Mean formulation.}
When the maximization in \eqref{eq:d-max} does not have a closed-form solution, fitting $\phi$ implies solving a min-max optimization problem.
This is potentially hard, arguably explaining the instability observed when training GANs \cite{GAN}.\footnote{%
    For example, if one considers $\mu_1$ to be the uniform measure on $(0, 1/4)\cup (3/4, 1)$, $\cF$ to be all the functions taking values in $[0,1]$, and $\phi$ to parameterize the uniform distributions on $(a, a+1/2)$ for $a\in[0,1/2]$, there is no stable saddle point that the optimization can converge too.
    Indeed, if the adversary is $f=\ind{(1/2, 1)}$, it will push $\phi$ to learn the uniform distribution on $(1/2, 1)$, which will lead the adversary $f$ to become $\ind{(0, 1/2)}$, which in turn will push $\phi$ to the uniform distribution there, creating an infinite loop.
}
Interestingly, one could equally consider some average with respect to a measure $\tau$ on $\cF$,
\begin{equation}
\label{eq:d-average}
    D_2^2(\mu, \mu_1; \tau) := \int_{\cF} \bscap{f}{\mu-\mu_1}^2\tau (\diff f).
\end{equation}
This integrand can be rewritten as 
\(
	\bscap{f}{\mu}^2 - 2\bscap{f}{\mu} \bscap{f}{\mu_1} + \bscap{f}{\mu_1}^2,
\)
which allows us to approximate the objective $D_2^2(\mu, \mu_1; \tau)$ empirically without bias thanks to the formula
\[
	\bscap{f}{\mu}^2 = \E_{X_1, X_2\sim\mu}[f(X_1)f(X_2)].
\]
Once again, when the span of the support of $\tau$ is dense in $\cC_1^{0,1}$, this defines a metric on measures.

\paragraph{Characteristic function matching.}
From a theory perspective, it is natural to consider $\cF = \brace{f_t:x\mapsto e^{tx} \midvert t \in T}$ for $T$ a subset of $\C$.
Those spaces of functions relate to the moment-generating  function $t\mapsto \E[e^{tX}] = \sum_{k\in\N} t^k \E[X^k] / k!$ and the characteristic function $t \mapsto \E[e^{itX}]$.
Using analytical properties, it turns out that as soon as $T \in \C^\N$ has a limit point, $\cF = \brace{f_t \midvert t \in T}$ can be used to define a metric through \eqref{eq:d-max} or \eqref{eq:d-average}.
It does not seems that those spaces have been utilized for generative AI, although we note that the specific case of $D_2$ where $\cF = \R\cdot i$ and $\tau(\diff f_{i\cdot s}) = \abs{s}^{-(d+1)} \diff s$ is known in the literature as the ``energy distance'',\footnote{%
    Seeing the characteristic function as the Fourier transform of a measure, it identifies with a negative Sobolev norm $H^{-d-1}$.
}
and can be rewritten as a MMD distance with $k(x, y) = \norm{x - y}$ \cite{ES}.

\paragraph{Generalization.}
While we have compared the same statistics on $\mu$ and $\mu_1$, one can generalize $D_\infty$ to $\cJ \subset L^1(\mu)\times L^1(\mu_1)$ under the form
\[
  D_\infty(\mu, \mu_1; \cJ) := \sup_{(f, g)\in \cJ} \bscap{f}{\mu} - \bscap{g}{\mu_1}.
\]
Taking the max over $(g, h) \in \cJ$ allows to cast $\phi$-divergences as $D_\infty$ for $\cJ = \brace{(f, \phi^*(g) | f:\X\to\R}$ where $\phi^*$ is the convex conjugate of $f$, as per Theorem 7.24 in \cite{fdiv}, see also \cite{rachev,Bottou2019}.
Moreover, this generalization describes Wasserstein distances thanks to Kantorovich duality,
\[
  W_c(\mu, \mu_1) := \min_{\pi \in \Pi_{\mu,\mu_1}}\E_{(x,y)\sim\pi}[c(x, y)]
  = D_\infty(\mu, \mu_1; \cJ_\Pi),
\]
where $c:\X\times\X\to\R_+$ is a cost function, $\Pi_{\mu, \mu_1}$ denotes all the probability measures on $\X\times\X$ whose marginals according to the first and second variables are $\mu$ and $\mu_1$ respectively, and $\cJ_\Pi$ is the set of pairs of function $(f, g)$ such that $f(x) - g(y) \leq c(x, y)$ for all $x, y \in \X$ \cite{Villani}.

\begin{figure}
    \centering
    \includegraphics{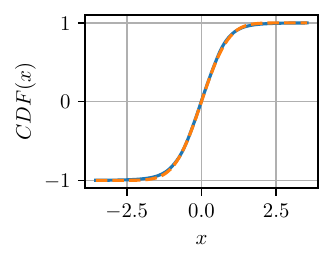}
    \vspace{-1.5em}
		\caption{Learning to map the Gaussian distribution to the uniform one with the mean formulation $D_2^2$, $\cF$ taken as the set of function $\brace{x\in[0,1]\mapsto e^{2\pi i kx} | k\in[10]}$, with $\tau$ uniform over those ten functions, and $\phi$ parameterized with a small multi-layer perceptron.
    A solution to this problem is given by the cumulative distribution function of the Gaussian, which relates to the error function.
    The ground truth is plotted in blue, while the learned pushforward is plotted in dashed orange.}
    \vspace{-1.5em}
    \label{fig:stat-matching}
\end{figure}

\section{Working at density level}\label{sec:density}

The second approach we will consider consists in maximizing the likeliness of a probability model.

When it admits a density, learning the distribution $\mu_1$ can be done through density estimation.
Once the density is learned, it can be used to generate new samples with techniques such as Monte Carlo Markov Chain.
Popular methods that go down this path are contrastive divergence and score matching, which learned the logarithm of the density based on Kullback-Leibler and Fisher divergence respectively together with computational tricks \cite{EBM}.
However, this work is focused on learning pushforward maps.

\paragraph{Trackable pushforward density.}
In particular the map $\psi:\X\to\X$, that parametrizes $\mu = \psi_\#\mu_0$ from some base measure $\mu_0 \in \prob\X$,\footnote{
    This parameterization ensures the Jacobian $D_z\psi$ to be a square matrix. If one uses $\mu = \phi_\#\nu$ with $\phi:\cZ\to\X$, then one would have to consider the restriction of $\phi$ on the image of $D_z\phi$, and $\diff x$ to be a base measure on this subset to write the change of variable formula.
} can be learned using the change of variable formula
\begin{equation}
    \label{eq:change-var}
	p(x) = p_0(\psi^{-1}(x)) \abs{\det\paren{D\psi^{-1}(x)}}.
\end{equation}
and plugging $p(x)$ into some variational objective.
Here, the densities $p$ and $p_0$ formally corresponds to the Radon-Nikodym derivatives of $\mu$ and $\mu_0$ respectively against a base measure, which is equally used to compute the Jacobian $D\psi$ of $\psi$,
\[
    p(x) := \frac{\diff\mu}{\diff x}(x),\quad p_0(x) := \frac{\diff\mu_0}{\diff x}(x).
\]
Once the map $\psi$ is learned, new samples are easily obtained as $\psi(X_0)$ for $X_0\sim\mu_0$ (supposedly easy to generate).
Following this perspective, \cite{push-density} suggests an invertible architecture with easily computable Jacobian determinants to be trained with log-likelihood maximization.
Composing many such mapping assimilates to a (normalizing) flow \cite{normal-flow}, which in the limit of composing infinitely many infinitesimal displacements can be seen as the solution of an ordinary differential equation \cite{NODE}.

\paragraph{Variational auto-encoder.}
Rather than providing new samples, $\phi(z)$ could be used as the parameter of a well-known distribution from which to sample $X$, e.g., $\paren{X | Z=z}$ is the Gaussian $\cN(\phi(z), 1)$.
This creates a joint probability model on $(X, Z)$ where the density $p(z)$ is fixed according to $\nu$, and $p(x|z)$ is to be optimized to maximize the likelihood of the data.
Computing the likelihood $p(x)$ implies a marginalization over $z$, which is usually too costly to compute.
However, the evidence lower bound (ELBO), stating that\footnote{%
    This statement follows from that fact that for an arbitrary $q_x\in\prob\cZ$, $\log p(x) = \E_{z\sim q_x}[\log(\frac{p(x, z)}{p(z|x)})] = \E_{z}[\log\paren{\frac{p(x, z)}{q_x(z)}}] + D_{KL}(q_x\| p(z|x))$,
    The maximum is reached for $q_x = p(z|x)$.
}
\[
  \log p(x) = \max_{q_x\in\prob\cZ} \E_{Z\sim q_x}[\log\paren{\frac{p(x, Z)}{q_x(Z)}}],
\]
allows to derive a tractable objective to learn $\phi:\cZ\to\X$ by maximizing $\sum_{i} \E_{Z\sim q_{x_i}}[\log p(x_i, Z) - \log q_{x_i}(Z)]$ over both $q:\X\to\prob\cZ$, seen as the encoder, and $\phi$ (which parameterizes $p(x|z)$), seen as the decoder \cite{VAE} (see also \cite{IW-VAE}).

\section{Building maps from flows}\label{sec:flow}
The third approach we will consider focuses on flows $t\mapsto\mu_t$ that transform continuously (with respect to some Wasserstein topology) the base measure $\mu_0$ into the target measure $\mu_1$. 

\paragraph{Stochastic differential equations.}
In order to build a map that goes from a Gaussian variable $\mu_0=\cN(0,I)$ to $\mu_1$, one could use a whitening process that map $\mu_1$ to $\mu_0$ before trying to invert it.
Stochastic processes offer such constructions.
E.g., one may initialize a random particle at $X_0\sim\mu_1$, and diffuse it according to an energy potential $U_t(x) = \nabla E_t(x)$ that pulls the particle towards low energy regions, together with random noise that excites the particle in various random directions.
Formally, the particle $X_t$ is initialized at $X_0\sim\mu_1$ and follows the stochastic differential process 
\begin{equation}
    \label{eq:sde}
    \diff X_t = - U_t(X_t) \diff t + \sqrt{2} \diff B_t,
\end{equation}
where $B_t$ is the Brownian motion.\footnote{%
    Inspired by physical systems, one could enrich equation \eqref{eq:sde} to describe particle interactions, see e.g. \cite{Coulomb}.
    While it has not been explored much in the literature, it may help to engineer better flow, or to enforce sample diversities.
}
The specific case where $E_t(x) = \norm{x}^2 / 2$ is known as the Ornstein-Uhlenbeck process, which maps $X_0\sim\mu_1$ to $X_\infty \sim \mu_0 = \cN(0, I)$, and could be reversed to go from $Z\sim\mu_0$ to $X\sim\mu_1$ through {\em stochastic simulation}.
This is the basis of diffusion models \cite{DPM}, but how does it relate to the original problem of finding a {\em deterministic mapping} between $\mu_0$ and $\mu_1$?
The answer leverages the flow $t\mapsto\mu_t$, defined through $\mu_t$ the law of $X_{(1-t)^{-1}}$.

\paragraph{From stochastic particle evolution to transport map.}
When a path that maps $t\in[0,1]$ to $\rho_t\in\prob\X$ happens to be absolutely continuous with respect to the $p$-Wasserstein topology, Theorem 8.3.1 of \cite{FlowMap0} implies the existence of a velocity field $v_t \in L^p(\rho_t)$ that governs the evolution of $\rho_t$ through the ``continuity equation''
\begin{equation}
    \label{eq:continuity}
    \diff \rho_t = -\Div(v_t \rho_t) \diff t.
\end{equation}
When $\rho_t$ is the law of $X_t$ that derives from equation \eqref{eq:sde}, $v$ is characterized by $v_t(x) = -U_t(x) - \nabla \log r_t(x)$, where $r_t(x)$ denotes the Lebesgue density of $\rho$ in $x$.
In essence, $v_t(x)$ characterizes the expected velocity of $X_t$ when at $X_t = x$.
Moreover, when $(t, x) \mapsto v_t(x)$ is bounded, Lipschitz continuous with respect to $x$, and continuous in $t$, Theorem 4.4 of \cite{OT} implies that the continuity equation \eqref{eq:continuity} can be integrated with $\chi_t:\X\to\X$
\begin{equation}
    \label{eq:ode}
    \diff \chi_t = v_t \diff t,
\end{equation}
from $\chi_0$ being the identity to obtain a mapping $\chi = \chi_1$ such that $\chi_\#\rho_0 = \rho_1$.\footnote{%
  Interestingly, many flow maps provide optimal transport plans \cite{OT-DDPM}.
}

\paragraph{The need to speed up inference.}
Recently, practitioners have focused on neural networks that map $(t, x)$ to $v_t(x)$. 
Such a network can be learned through score or flow matching \cite{flow-match}.
They then use it at inference time to simulate the reverse stochastic process \eqref{eq:sde} leading from $Z\sim\mu_0$ to $X\sim\mu_1$, or to compute trajectories $t\mapsto\chi_{(1-t)^{-1}}(Z)$ by integration of the ODE \eqref{eq:ode}.
While diffusion models do not seem to be particularly more principled than other approaches, they are performing remarkably well on images in practice.
This may be because $(t,x)\mapsto v_t(x)$ is expected to be regular, hence easy to learn, especially with some adapted U-net architecture; and because the error between trajectories obtained by integration of an estimated flow $\hat v_t$ and integration of the real one $v_t$ is relatively well-behaved (see, e.g., Section 2.4 in \cite{interpolant}).

However, generating new samples with flow-based models currently relies on expensive stochastic simulation or flow integration at inference time.
Some attempts have been made to learn straight flows that are fast to integrate \cite{rec-flow}.
Others have tried to globally integrate the velocity $v_t$ into a map $\psi$, which has been referred to as distillation, as one want to ``distill'' the neural network that parameterized $(t, x) \mapsto v_t(x)$ into a new neural network parameterizing $\psi$ \cite{distillation}.
But why not directly learn the pushforward $\psi = \chi_\infty^{-1}$ rather than going through the convoluted construction of learning and integration of flows?

\paragraph{From network to optimization architecture.}
Let us now focus on the following research question:
GANs are known to be hard and unstable to train, but could diffusion models help us engineer better optimization procedures?
Before training, a GAN (or an auto-encoder) will more or less map Gaussian noise, seen as the distribution $\mu_0$, to Gaussian noise, seen as the distribution $\rho_\infty$.
Eventually, one could fine-tune it iteratively to map $\mu_0$ to $\rho_t$ for $t$ decreasing during the different rounds of fine tuning.
At the end, one would obtain a mapping from $\mu_0 = \rho_\infty$ to $\mu_1 = \rho_0$.
The fine-tuning would always employ the same loss at each iteration, but the data would be modified little by little, starting with really noisy images, i.e., $X \sim \rho_t$ for a big $t$, to sharper high-resolution images, i.e., $X\sim\rho_t$ for a small $t$.

Diffusion models have seduced theorists, they may wonder what our suggestion corresponds to in more abstract terms.
Extending on the seminal papers of Otto \cite{ODEWasserstein0,ODEWasserstein1}, one can show that the flow $t\mapsto\rho_t$ that solves \eqref{eq:sde} for a time-independent potential $U_t = \nabla E$ corresponds to the gradient flow (for some topology) of the objective $D_2$ \eqref{eq:d-average} for $\cF = \brace{f_i \vert i\in\N}$ and $\tau(f_i) \propto \lambda_i$, where $(\lambda_i, f_i)$ is the eigendecomposition of the operator representing the quadratic form $f\mapsto \bscap{\norm{\nabla f}^2}{\mu_0}$ in $L^2(\mu_1)$ where $\mu_1 \propto \exp(-E(x))\diff x$.\footnote{%
    It is relatively standard to show that $\diff\bscap{f_i}{\rho_t} = -\lambda_i \bscap{f_i}{\rho_t}\diff t$ for all $i\in\N$ and that $f_0 = 1$ \cite{Diffusion}. 
    As a consequence, $\bscap{f_i}{\rho_\infty} = \bscap{f_i}{\mu_0} = 0$, and 
    \(
        \frac{\diff \rho_t}{\diff t} = -\sum_i \lambda_i \bscap{f_i}{\rho_t}f_i^* = -\frac{\diff}{\diff \rho} \sum_i \lambda_i \bscap{f_i}{\rho_t}^2,
    \)
    which assimilates to $\diff D_2(\rho_t, \mu_0; \tau) / \diff \rho$.
    Details to define the different topologies (to take adjoints and gradients) and prove formally this result are beyond the scope of this paper.
}
In other terms, denoising diffusion probability models that learn $v_t$ through the Ornstein-Uhlenbeck process so that the base sampling distribution $\mu_0(\diff x) \propto \exp(-x^2)$ is a Gaussian distribution are implicitly following and reverting a gradient flow.
If diffusion models encode the full flow $(t, x) \mapsto v_t(x)$ \eqref{eq:continuity} in their architecture, we are suggesting to rather encode this flow into an optimization scheme to learn a network $f_\theta$ so that after $t$ ``epochs'' of optimization $f_{\theta_{t}} = \psi_{t}^{-1}$ \eqref{eq:ode}, and at the end of the optimization $(f_{\theta_\infty})_\#\mu_0 = \mu_1$.

\section{Conclusion}
There exists many perspectives on sampling, even when we restrict ourselves to the search of a pushforward map while only accessing samples from the target distribution.

A straightforward approach consists in defining losses using test functions to quantify differences between measures, as explained in Section \ref{sec:functions_matching}.
Among the methods based on this principle, GANs have provided the most exciting results, but their training is hard, requiring skilled engineering.

This has led to the rise of another stream of research based on density-centered considerations, as detailed in Section \ref{sec:density}.
One can e.g. learn pushforward maps by maximizing probability likelihood, though it involves ensuring mapping invertibility and computing Jacobian determinants, which is constraining.
Moreover, the target distribution $\mu_1$ might not have a density, which is notably the case when the data span uniformly a manifold in the input space.
This hurdle can be avoided by artificially injecting noise to the data to create a distribution $\mu_1'$ admitting a density before adding a denoising step to recover $\mu_1$ from $\mu_1'$.
Several model evolutions have led to cascading many noise injection and denoising processes, which were ultimately understood through the lens of Brownian motion and reversible diffusion models, explained in Section \ref{sec:flow}.
Those diffusion models are currently state-of-the-art, although they require extensive simulation to generate new samples.
Remarkably, they are implicitly associated with pushforward maps.
 
Getting a neural network to efficiently learn such a pushforward map would be a major advance for generative AI.
Another, perhaps less discussed issue is that the target measure is only accessed through a finite number of samples, and rules to extrapolate from them arise mainly as a consequence of the biases induced by network architectures and optimization schemes--more principledness would be desirable.

This paper reviewed several principles to learn pushforward maps.
Future work consists in experimenting with those to extract more practical knowledge for generative AI across different modalities.

\paragraph{Acknowledgment.}
VC would like to thank Brandom Amos, Alberto Bietti, Florian Bordes, L\'eon Bottou, Ricky Chen, Valentin de Bortoli, Carles Domingo-Enrich, Yann LeCun, Loucas Pillaud-Vivien, Aram Pooladian and Yann Olivier for useful discussions.

\bibliographystyle{template/IEEE}
\bibliography{reference}

\end{document}